\documentclass[journal] {IEEEtran}
\usepackage{floatrow}
\DeclareFloatFont{normalsize}{\normalsize}
\usepackage{amsmath}
\usepackage{graphicx}%
\usepackage{amsfonts}%
\usepackage{amssymb}
\usepackage{amsthm}  %
\usepackage{subfig}
\usepackage{tabularx}
\usepackage{multirow}
\usepackage{makecell}
\usepackage{adjustbox}

\usepackage{caption}

\title{\huge Human Action Recognition Using Deep Multilevel Multimodal ($M^2$) Fusion of Depth and Inertial Sensors}
\author{Zeeshan Ahmad, \textit{Student Member, IEEE}, Naimul Khan, \textit{Member, IEEE}}

\begin{document}
	\maketitle
	\begin{abstract}
Multimodal fusion frameworks for Human Action Recognition (HAR) using depth and inertial sensor data have been proposed over the  years. In most of the existing works, fusion is performed at a single level (feature level or decision level), missing the opportunity to fuse rich mid-level features necessary for better classification. To address this shortcoming, in this paper, we propose three novel deep multilevel multimodal ($M^2$) fusion frameworks to capitalize on different fusion strategies at various stages and to leverage the superiority of multilevel fusion. At input, we transform the depth data into depth images called sequential front view images (SFIs) and inertial sensor data into signal images. Each input modality, depth and inertial, is further made multimodal by taking convolution with the Prewitt filter. Creating ``modality within modality'' enables further complementary and discriminative feature extraction through Convolutional Neural Networks (CNNs). CNNs are trained on input images of each modality to learn low-level, high-level and complex features. Learned features are extracted and fused at different stages of the proposed frameworks to combine discriminative and complementary information. These highly informative features are served as input to a multi-class Support Vector Machine (SVM). We evaluate the proposed frameworks on three publicly available multimodal HAR datasets, namely, UTD Multimodal Human Action Dataset (MHAD), Berkeley MHAD, and UTD-MHAD
Kinect V2. Experimental results show the supremacy of the proposed fusion frameworks over existing methods.

	\end{abstract} 

\begin{IEEEkeywords}
 Canonical correlation analysis, fusion of depth and inertial sensors, human action recognition,, multimodal fusion.
	\end{IEEEkeywords}		

	\section{Introduction}
	
\IEEEPARstart{H}{UMAN} Action Recognition (HAR) has been a core area of research for the multimedia community due to its applications in various fields, including human computer interaction~\cite{chen2014medication}, healthcare~\cite{corbishley2008breathing}, sports~\cite{zhou2016never},visual surveillance~\cite{qin2016compressive},  military, robotics and gaming.\let\thefootnote\relax\footnote{© 2019 IEEE. Personal use of this material is permitted. Permission from IEEE must be obtained for all other uses, in any current or future media, including reprinting/republishing this material for advertising or promotional purposes, creating new collective works, for resale or redistribution to servers or lists, or reuse of any copyrighted component of this work in other works.}

Before the resurgence of neural networks, especially the deep learning models, conventional methods for HAR were statistical methods, where the focus was on designing hand-crafted features. These hand-crafted methods have two disadvantages: 1) requirement of domain knowledge about the data~\cite{plotz2011feature}; 2) capturing only a subset of the features; resulting in difficulty to generalize for unseen data. The exemplary performance of deep learning  in classification tasks~\cite{krizhevsky2012imagenet} has diverted the attention of researchers from hand-crafted methods to deep learning models. CNN in particular has gained significant attention in computer vision and machine learning due to its strong ability to automatically learn invariant and hierarchical features directly from the images.

Low cost sensors including Kinect depth cameras, wearable sensors and availability of smartphones have been very beneficial for HAR. Kinect depth cameras provide 3D action data, require less hardware resources and are less sensitive to lighting changes and clutter as compared to RGB cameras~\cite{aggarwal2014human}. However there are limitations associated with depth images like view point variation, noise during image acquisition and constrained space defined by camera orientation. These shortcomings can be alleviated by utilizing wearable inertial sensor for HAR such as accelerometer and gyroscope~\cite{yang2009distributed}. Wearable sensors provide multivariate time series data in terms of 3-axis accelerations from accelerometers and three axis angular velocities from gyroscopes. These sensors provide data at a high sampling rate and can work in dark and unbounded environment. Like depth cameras, wearable sensors have limitations such as sensor drift and intrusiveness, as the humans have to physically wear them~\cite{chen2017survey}. 

Multimodal fusion can alleviate the shortcomings of both types of sensors, and thus improve the performance of HAR. Although multimodal HAR has progressed significantly over the past few years, the current works fall short of optimal performance when combining multiple modalities. A major difficulty is in deciding at which level of information fusion: early fusion, feature level fusion and decision/late fusion, the modalities should be fused~\cite{ramachandram2017deep}.  

The purpose of multimodal fusion is to obtain complementary information from modalities to perform the analysis task accurately. In multimodal fusion the major consideration
is to find the optimal instance or stage to fuse the modalities. Based on this philosophy,
the commonly used strategies are data level or early fusion, feature level or intermediate
fusion and decision level or late fusion~\cite{hall1997introduction}. The emergence of deep learning has resulted in some new concepts for fusion. Since feature extraction in deep learning models, especially the CNN, can be performed at any layer, feature level fusion for deep networks can be further divided into early feature level fusion (fusion between features of convolutional layers) and late feature level fusion (fusion between features of fully connected layers) depending upon the layers of the deep network from which the features are extracted.

Feature level fusion is the most widely used approach for integrating information in
deep learning models. The greatest benefit of feature level fusion is that it utilizes the
correlation among the modalities at an early stage.
Furthermore, only one classifier is required to perform a task, making the training process
less tedious. However, a notable limit of feature level
fusion is time synchronization, since the data in various modalities are captured at
different rates and formats~\cite{wu2006multi}.
The other broadly used fusion tactic is decision level. The significant advantage of this
technique is that it allows us to explicitly examine each modality and thus the chance of
dominance of one modality over the other is greatly reduced. Furthermore, decision level
fusion offers scalability in terms of modality~\cite{atrey2007goal}, since adding a new modality is relatively easier. The notable issue with decision
level fusion is the use of more than one classifier. This makes the task time
consuming. Hybrid multimodal fusion has been exercised by investigators to cop up the
shortcomings of feature level and decision level fusion~\cite{ni2004image}. However,  there is no single fusion method which can be considered as an ultimate solution yet. Most current works either do feature level fusion or decision fusion, missing the opportunity of fusing rich mid-level feature representations that are available in a CNN-based architecture.

To address the aforementioned deficiencies and to exploit different fusion strategies, in this paper, we present three novel deep
multilevel multimodal ($M^2$) fusion frameworks for HAR. In each distinct fusion framework, we applied multilevel fusion to counter the short comings of single stage fusion so that an optimal framework can be reached methodically. The key contributions of the presented work are:
 \begin{enumerate}	
 	
 		\item We propose three unique deep $M^2$ fusion frameworks : ``Deep multistage feature fusion framework'', ``Deep hybrid fusion framework'' and ``Computationally efficient fusion framework''. The purpose of presenting three deep fusion frameworks is to get full advantage of multimodal fusion by applying multiple fusion strategies to find a robust framework for HAR. We show that through careful execution of multilevel multimodal ($M^2$) fusion, all three fusion frameworks can be tuned to provide outstanding results for depth-inertial HAR. 
 			
 	\item Inspired from the outstanding performance of CNN on image classification task~\cite{krizhevsky2012imagenet}, we transformed both input datasets, depth and inertial, into depth and signal images respectively. To extract more discriminative and complementary features, we increased the number of modalities by taking convolution of SFIs and signal images with the Prewitt filter. Converting input data to images and generating additional modality by Prewitt filtering enables extraction of different feature types (e.g. edges, curves, and higher-level abstractions) with CNN that are not possible with 1D temporal data~\cite{hatami2018classification}.

   \end{enumerate}	
The rest of the paper is organized as follows. Section II describes the related works on HAR using conventional and deep learning-based fusion. Section III provides technical details of our signal processing and image conversion ideas, and the proposed deep $M^2$ fusion frameworks. In Section IV, we provide detailed experimental analysis, where the two aforementioned contributions are analyzed in detail through ablation studies and comparison with state-of-the-art models. Section V concludes the paper.

\section{Related Work}

Since HAR is a very active research field with many research directions, for brevity, we only discuss existing methods for HAR that attempts at fusing different modalities.

In~\cite{chen2015improving} the accuracy of HAR is improved by fusing features extracted from depth and inertial sensor data and using collaborative representation classifier. Improved accuracy results were achieved due to complementary aspect of data from both modalities. A decision level fusion is performed between depth camera data and wearable sensor data to increase the capabilities of robots to recognize human actions in~\cite{manzi2018enhancing}. An efficient real-time human action recognition system is developed in~\cite{chen2016real} using decision level fusion of depth and inertial sensor data. Depth and inertial data is effectively merged in~\cite{liu2014fusion} to train a hidden Markov model for improving accuracy and robustness of hand gesture recognition.  In~\cite{dawar2018real}, a computationally efficient real-time detection and recognition approach is presented to identify actions in the smart TV application from continuous action streams using continuous integration of information obtained from depth and inertial sensor data. A method for bilateral gait segmentation is proposed in~\cite{hu2018novel} by multimodal fusion of features obtained from thigh mounted inertial sensor  and depth sensor with the contralateral leg in its field of view. The proposed method can be used to make lower limb assistive devices for patients with walking impairments. In~\cite{guo2017multiview},  novel and robust unsupervised method based on fusion of depth and inertial sensor data for HAR is proposed. The proposed Multiview Cauchy Estimator Feature Embedding (MCEFE) method is capable of finding the optimal unified space and projection matrices by minimizing empirical risk through the Cauchy estimator. A comprehensive survey on fusion of depth and inertial sensors is provided in~\cite{chen2017survey} where the recent success of the fusion and future challenges and trends are discussed in details.

	Due to the recent popularity of deep neural networks (DNN) in multimedia applications, several deep learning based fusion frameworks for HAR has recently been presented.  In~\cite{bernal2018deep} a supervised deep multimodal fusion framework for process monitoring and verification in the medical and healthcare fields is presented that depends on simultaneous processing of motion data acquired with wearable sensors and video data acquired with body-mounted camera. Authors in~\cite{hwang2017multi} proposed DNN based fusion of images and inertial data for improving the performance of human action recognition. Two CNNs were used to extract features from images and inertial sensors and the fused fearures were used to train an RNN classifier. 
	
	Deep learning based fusion methods for HAR using depth and inertial sensors have been presented in~\cite{dawar2019data},~\cite{dawar2018action},~\cite{ahmad2018towards} and~\cite{dawar2018convolutional}. In~\cite{dawar2019data}, CNN is used to extract features from depth images while recurrent neural netwrok (RNN) is used to capture features from inertial sensor data. Finally, a decision level fusion is performed on extracted features to improve the accuracy of HAR. In~\cite{dawar2018action} deep learning based fusion system based on fusing the depth and inertial data is presented which is capable of detecting and recognizing actions of interest from continuous action streams. CNN is used to extract features from depth images, and a combination of CNN
	and long short-term memory (LSTM) network is utilized for inertial
	signals. First, the segmentation is performed on each sensing modality and then actions of interest were detected. Finally, the decision level fusion is performed for recognition. In~\cite{ahmad2018towards}, CNNs are used to extract features from depth images obtained from depth sensor and from signal images obtained from inertial data. Finally, a concatenation fusion is performed between the two modalities. A CNN based sensor fusion system is developed in~\cite{dawar2018convolutional} to detect and monitor transition movements between body states as well as falls in healthcare applications. Fusion between the modalities is carried out by accumulating the scores of fully connected layers of CNNs. 
	
	The major shortcoming in existing deep learning based fusion methods for HAR using depth and inertial sensors is that the fusion is performed at a single level or stage, either feature level or decision level, thus failing to map the true semantic information from data to classifier. Since deep learning models allow us to extract features at all levels of their structure, we get rich multilevel features comprised of low level, mid-level and high level features. The existing methods do not take advantage of this rich multi-level information, leaving room for improvement in the state-of-the-art of deep learning-based visual-inertial action recognition.
		
	 To adress all the shortcomings of the related work, in this paper, we introduce three novel deep multilevel multimodal ($M^2$) fusion frameworks for improving the performance of HAR using depth and inertial sensor data. In all three fusion frameworks, fusion is performed at multiple stages to alleviate the deficiencies of the existing methods. The three ($M^2$) fusion frameworks are explained in detail in section~\ref{proposed method}.
	
	The proposed frameworks are an extension of our recent work~\cite{ahmad2018towards}, where a CNN-based fusion framework is presented, depth and inertial sensor data is transformed into images and then CNNs are employed to extract features from the transformed images. The depth and image features are simply concatenated (feature level fusion) and served as input to train a multiclass SVM classifier. We extend the work here by proposing the novel $M^2$ fusion frameworks that perform multilevel multlimodal fusion. 
	
	\section{Proposed Method}\label{proposed method}
	In this section, we first describe the common components of our frameworks; converting input signals to images, creating modality within modality, and fusion technique. Then, the three novel deep multilevel multimodal ($M^2$) fusion frameworks are presented.

	\begin{figure}[h]
		\centering
		\includegraphics[width=0.9\linewidth]{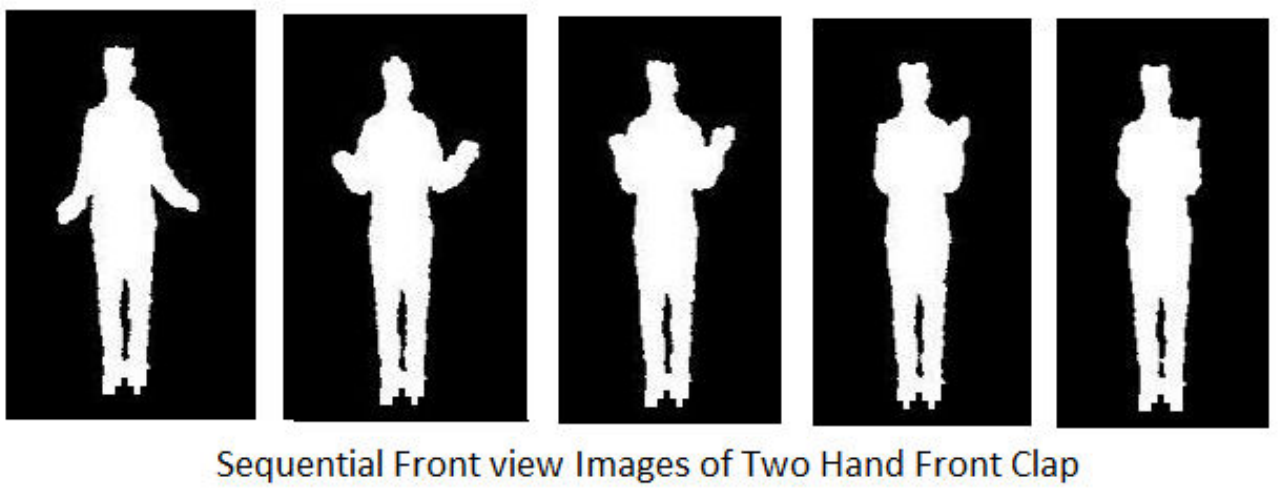}
		\caption{Five samples of Sequential Front View Images of Two hand Front Clap}
		\label{fig:sequential front view images}
	\end{figure}
\begin{figure}[h]
	\centering
	\includegraphics[width=0.9\linewidth]{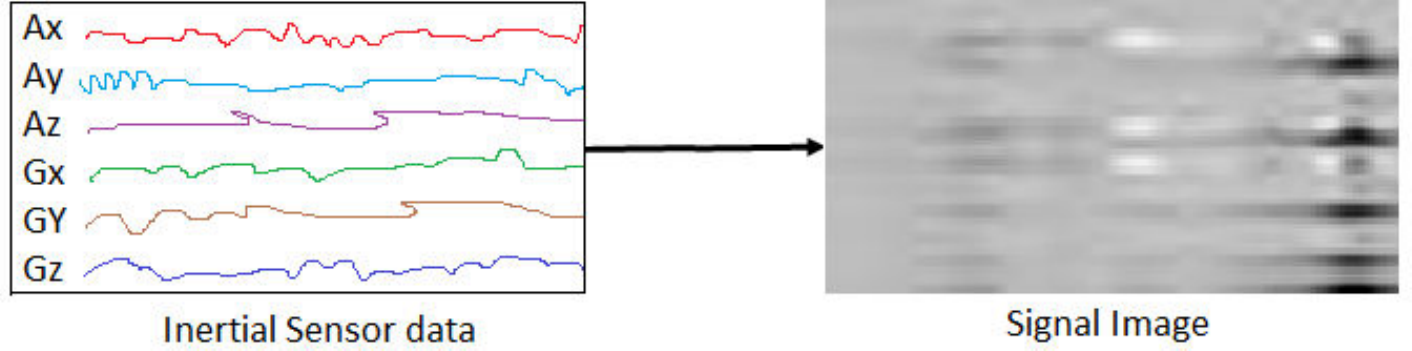}
	\caption{Formation of Signal Image from Inertial sensor data}
	\label{fig:conversion}
\end{figure}	
\begin{figure}[h]
	\centering
	\includegraphics[width=0.9\linewidth]{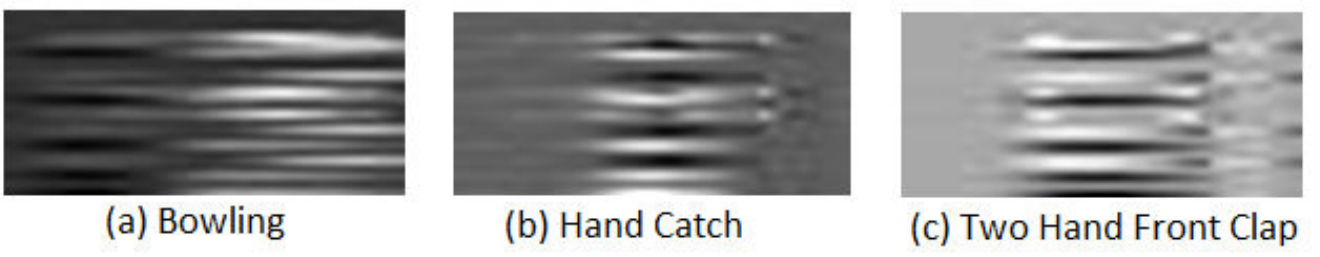}
	\caption{Signal Images of three different actions}
	\label{fig:signal images}
\end{figure}

	\subsection{Formation of Sequential Front View Images}\label{SFIs}

The depth data provides 3D information, therefore we can generate  the front, top and side view of depth motion maps. We experimentally observe that only front view information is enough to recognize the actions as the fusion of front view with other views doesnot significantly increase the recognition accuracy as shown in Table~\ref{tab:Baseline}, but only increase the computational cost. The supplementary information is provided by the inertial dataset. Thus we convert the depth sequences into images called Sequential Front view Images (SFI) as shown in Fig.~\ref{fig:sequential front view images}. By using only one view in the SFIs, we are reducing the computational cost. These images are similar to the motion energy images and motion history images introduced in~\cite{bobick2001recognition}. These SFIs provide cumulative information about the action from start to completion.

\subsection{Formation of Signal Images}\label{signal images}
Inertial sensors generate data in the form of multivariate time series. In our datasets, we have six sequences of signals : three accelerometer and three angular velocity sequences obtained from accelerometers and gyroscopes respectively.

We converted six sequences into 2D virtual images called signal images based on the algorithm in~\cite{jiang2015human}. The conversion of time series data to signal image is shown in Fig.~\ref{fig:conversion}. Signal image is obtained through row-by-row stacking of given six signal sequences in such a way that each sequence appears alongside to every other sequence.  The signal images are formed by taking advantage of the temporal correlation among the signals. 

Row wise stacking of six sequences has the following order.

\textit{123456135246142536152616}

Where the numbers \textit{1} to \textit{6} represent the sequence numbers in a raw signal. Order of the sequences clearly shows that every sequence neighbors every other sequence to make a signal image. Thus the final width of signal image becomes 24. 

For deciding the length of the signal image we made use of sampling rate of datasets which is 50Hz for our two datasets. Therefore, to capture granular motion accurately, the length of the signal image is finalized as 52, resulting in a final image size of 24 x 52. These signal images are shown in Fig.~\ref{fig:signal images}.

\subsection{Modality within modality}

We create a modality within each input modality by convolving images of each input modality with the Prewitt filter. Prewitt filters are edge detectors and are simple to implement~\cite{gonzalez2002digital}. The filters deployed by CNN are trainable and are specifically tuned to a dataset through a long training process. The purpose of applying Prewitt filter was to create a generic additional modality for any dataset that can be created efficiently without training, before we feed multiple modalities to our proposed networks. This can be thought of as an additional step of pre-processing to create an additional modality. Our purpose is to show that creating an additional modality can help extracting complementary and discriminative features. While we used Prewitt filter to demonstrate our multimodal networks, we believe other type of filters can be utilized too. Creating edge oriented modalities for both depth and signal images using Prewitt filter have been experimentally proved significant as shown in Table~\ref{tab:UTDMHAD comparisonTabe}. These modalities allow CNN to extract further complementary and discriminative features and thus the availability of rich features due to these created modalities help classifiers to perform its task efficiently and accurately.

 We apply the following 3-by-3 Prewitt filter that emphasizes horizontal edges.
\[
h=
\begin{bmatrix}

1 & 1 & 1 \\
0 & 0 & 0 \\
-1 & -1 & -1

\end{bmatrix}
\]	

\begin{figure}[h]
	\centering
	\includegraphics[width=0.9\linewidth]{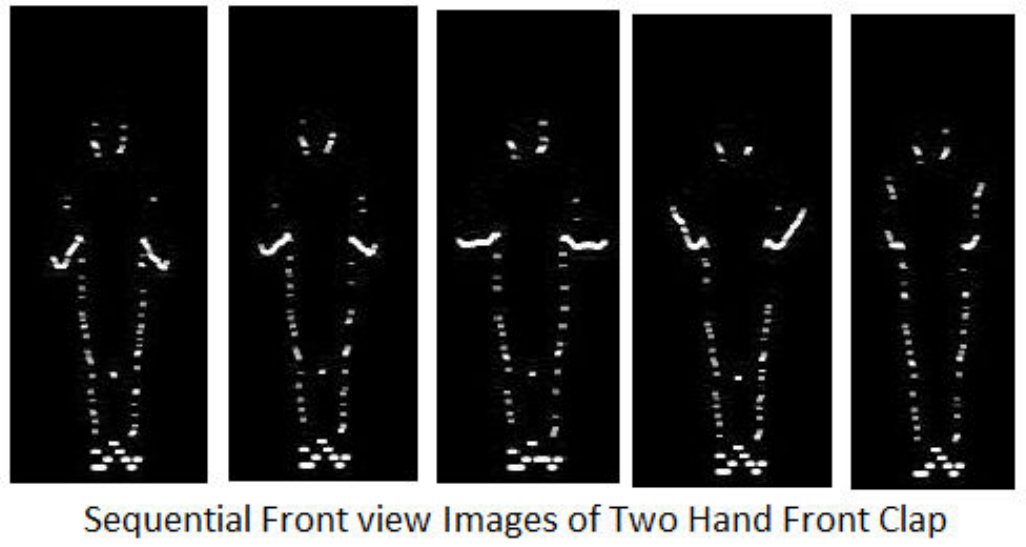}
	\caption{Prewitt Filtered Sequential Front view Images of Two Hand Front Clap}
	\label{fig:Prewitt Filter SFI}
\end{figure}

\begin{figure}[h]
	\centering
	\includegraphics[width=0.9\linewidth]{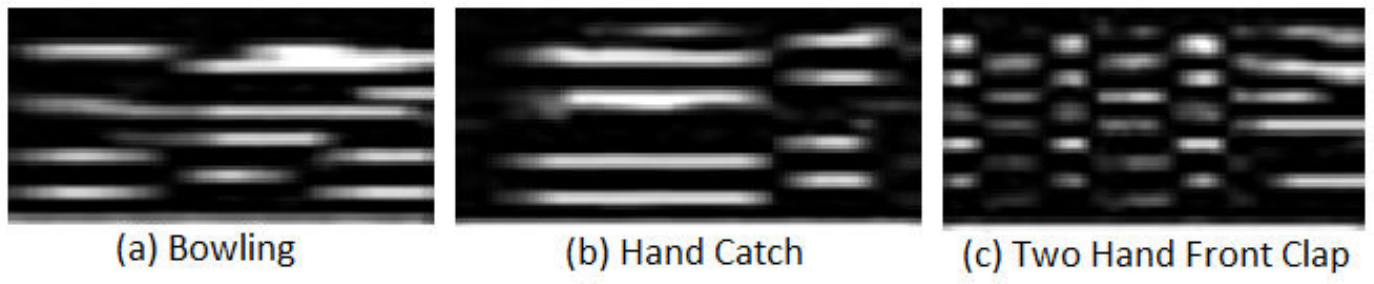}
	\caption{Prewitt Filtered Signal Images of three different actions}
	\label{fig:Prewitt Filter signal images}
\end{figure}

Prewitt filtered sequential front view images and signal images are shown in Fig.~\ref{fig:Prewitt Filter SFI} and~\ref{fig:Prewitt Filter signal images} respectively.

\subsection{Canonical Correlation based Fusion (CCF)}\label{CCA Fusion}
Canonical correlation analysis (CCA) is effective and robust multivariate statistical method for finding the relationship between two sets of variables.

Let $X\in\mathbb{R}^{p \times n}$ and $Y\in\mathbb{R}^{q \times n}$ represents the feature matrices from two modalities, where $p$ and $q$ are the dimensions of the first and second feature set respectively and $n$ are the training samples in each modality. let $\Sigma_{xx}\in \mathbb{R}^{p \times p}$ and $\Sigma_{yy}\in \mathbb{R}^{q \times q}$ denote the within set covariance matrices of $X$ and $Y$ respectively and $\Sigma_{xy}\in \mathbb{R}^{p \times q}$ 
denotes the between set covariance matrix for $X$ and $Y$ and $\Sigma_{yx}=\Sigma_{xy}^T$. The overall augmented covariance matrix of size $(p+q)\times(p+q)$ is given by

\begin{equation}\label{first correlation equation}
cov(X,Y) = 
\begin{bmatrix}

\Sigma_{xx} & \Sigma_{xy} \\
\Sigma_{yx} & \Sigma_{yy}

\end{bmatrix}
\end{equation}

The purpose of CCA is to find the linear combination $X^\prime=AX$ and $Y^\prime=BY$ such that the maximum pairwise correlation between the modalities could be achieved. Matrices $A$ and $B$ are called transformation matrices for $X$ and $Y$ respectively. The correlation between $X^\prime$ and $Y^\prime$ is given by  

\begin{equation}
corr(X^\prime,Y^\prime) = \frac{cov(X^\prime,Y^\prime)}{var(X^\prime).var(Y^\prime)}
\end{equation}
where $cov(X^\prime,Y^\prime)=A^T\Sigma_{xy}B$, $var(X^\prime)= A^T\Sigma_{xx}A$ and $var(Y^\prime)= B^T\Sigma_{yy}B$. $X^\prime$ and $Y^\prime$ are known as canonical variates.

Lagrange's Optimization method is used to maximize the covariance between $X^\prime$ and $Y^\prime$ subject to the constraint that the variance of $X^\prime$ and variance of $Y^\prime$ is equal to unity~\cite{uurtio2018tutorial}.
\[
var(X^\prime)=var(Y^\prime)=1 
\]

On the transformed feature vectors, canonical correlation based fusion (CCF) is performed by  adding the transformed feature vector. Mathematically this addition is written as 
\[
Z =  X^\prime + Y^\prime = A^T X + B^T Y \]

\begin{figure}
		\centering
		\includegraphics[width=\linewidth]{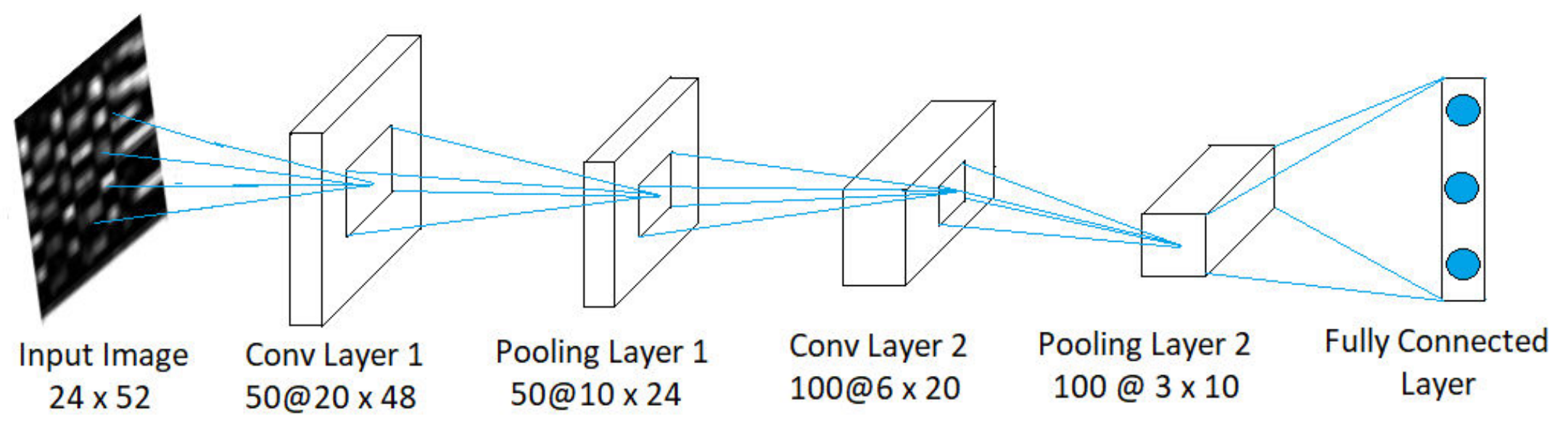}
		\caption{CNN Architecture for Signal Image. Consists of two convolutional layers, two pooling layers, and a fully connected layer. The first convolutional layer has 50 kernels of size 5x5, followed by pooling layer of size 2x2 and stride 2. The output of the first pooling layer is the input of the second convolutional layer which has 100 kernels followed by 2x2 pooling layer with stride 2.}
		\label{fig:CNN Architecture}
	\end{figure}

\vspace{-0.5cm}
\subsection{$M^2$ Fusion frameworks}	

Finally we present the fusion frameworks utilizing the aforementioned common components. 

\begin{figure*}
		\centering
		\includegraphics[width=0.7\linewidth]{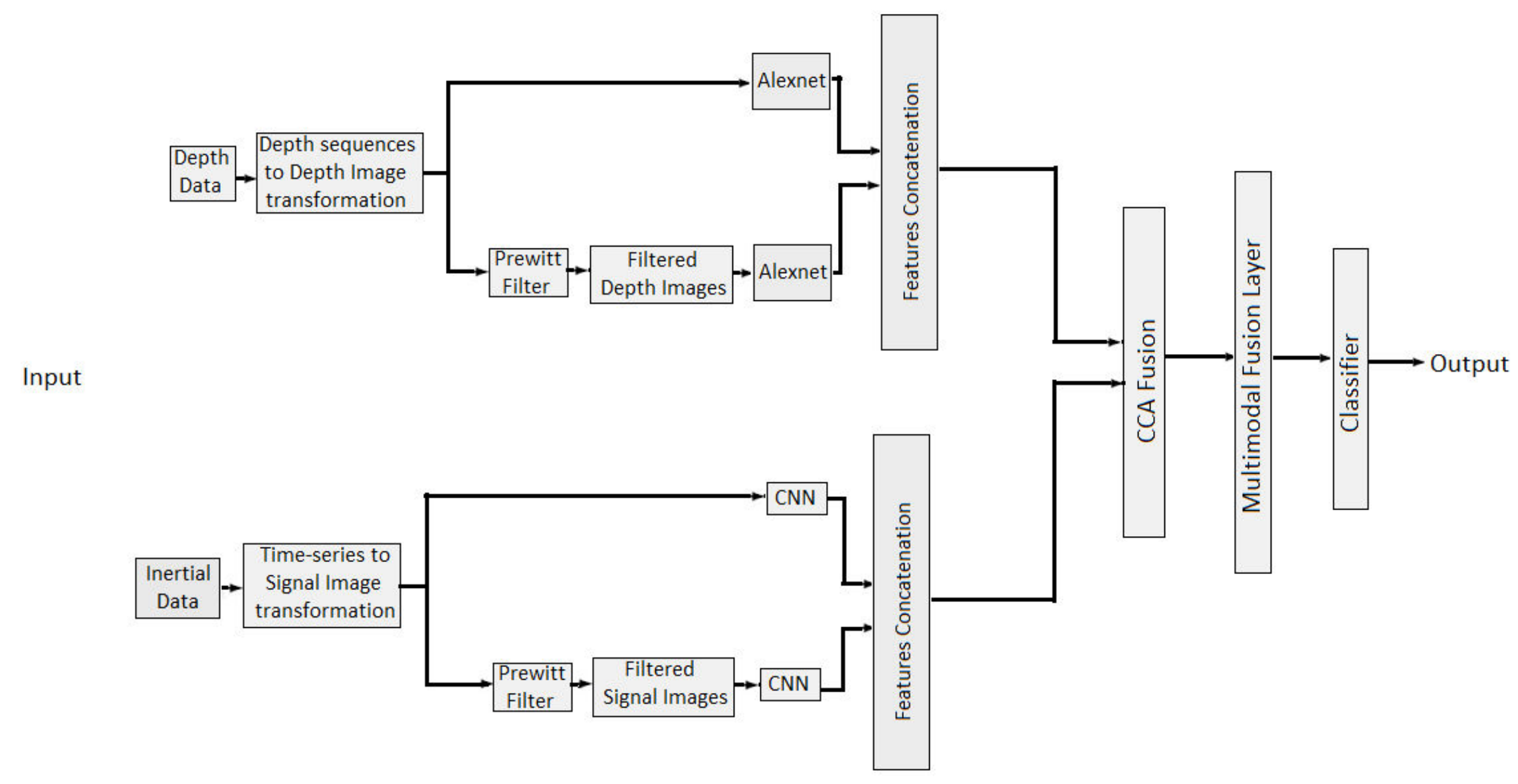}
		\caption{Deep Multistage Feature Fusion Framework}
		\label{fig:first framework}
	\end{figure*}
	\begin{figure*}
		\centering
		\includegraphics[width=0.7\linewidth]{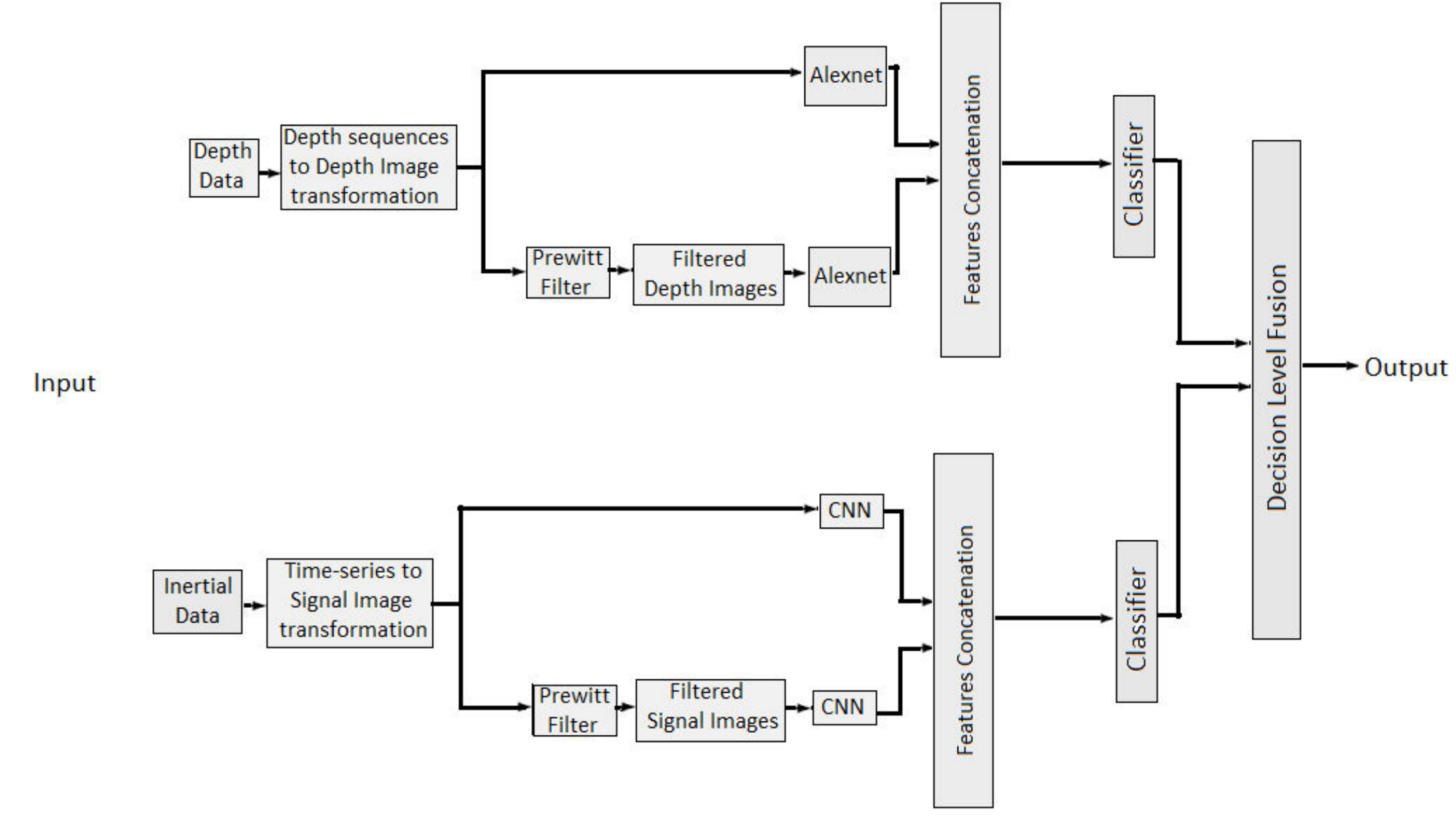}
		\caption{Deep Hybrid Fusion Framework}
		\label{fig:second framework}
	\end{figure*}
	\begin{figure*}
		\centering
		\includegraphics[width=0.7\linewidth]{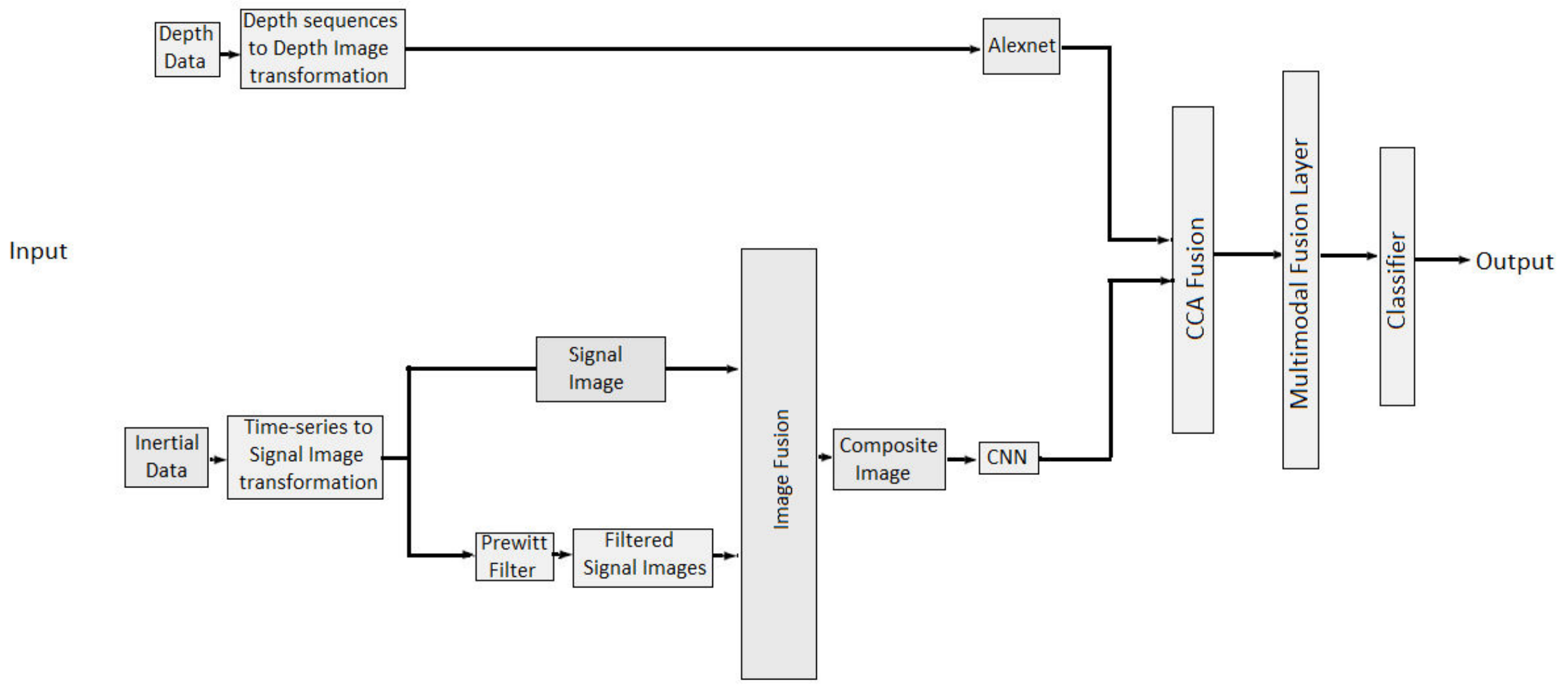}
		\caption{Computationally Efficient Fusion Framework.}
		\label{fig:third framework}
	\end{figure*}

	\subsubsection{Deep Multistage Feature Fusion Framework}\label{First Fusion framework}
	Our first network, the \textit{deep multistage feature fusion framework} is built upon the philosophy of processing the separate modalities through separate CNNs first, which act as feature extractors. Then, a two stage feature fusion is performed, feature concatenation, followed by canonical correlation based fusion (CCF). The architecture of framework is shown in Fig.~\ref{fig:first framework}.
	
	Alexnet (CNN based model)~\cite{krizhevsky2012imagenet} is used to extract features from SFIs and Prewitt filtered SFIs. Another smaller CNN, whose architecture is shown in Fig.~\ref{fig:CNN Architecture}, is used to extract features from signal images and Prewitt filtered signal images as shown in Fig.~\ref{fig:first framework}. We extracted learned features from the second last fully connected layer of both Alexnets and performed concatenation. Similarly, learned features extracted from the first fully connected layer of the CNNs for signal images are also concatenated. The final fusion with CCF is performed between the concatenated features of both modalities as shown in Fig.\ref{fig:first framework}. For CCF, we downsampled the concatenated features of depth modality using bicubic interpolation, so that they are the same size as the concatenated features of inertial modality to avoid rank deficiency during calculations of transformation matrices. Fusing the two concatenated layers using CCF results in highly discriminative features compared to simple concatenation~\cite{sun2005new}. This fused data is served as input to multiclass SVM for performing recognition task. 
	
	We used SVM as a classifier since we experimentally proved in our previous work~\cite{ahmad2018towards} that SVM performs better than softmax, which is typically built into any CNN framework. Softmax classifier reduces the crossentropy function while SVM employs a margin based function. Multiclass SVM classifies data by locating the hyperplane at a position where all data points are classified correctly. Thus SVM determines the maximum margin among the data points of various classes. The more rigorous nature of classification is likely the reason why SVM performs better than softmax. Since we are doing multilevel fusion, therefore we have to extract and fuse features at more than one stages and hence we require an external classifier, thus classification within the CNN framework is meaningless and  end-to-end deep learning model cannot be promoted.

\subsubsection{Deep Hybrid Fusion Framework}\label{second fusion framework}
The architecture of the \textit{deep hybrid fusion framework} is shown in Fig.~\ref{fig:second framework}.

This framework uses hybrid fusion, combination of feature level and decision level fusion, as compared to deep multistage fusion framework that used feature level fusion at two stages. Furthermore two classifiers are used to generate scores for each modality to perform decision level fusion. 

At the input of this framework, input modalities are converted into images and features are extracted with Alexnet and CNN and fused by concatenation in same way as that of multistage fusion framework explained in section~\ref{First Fusion framework}. The two concatenated feature vectors are separately inputted to two SVM classifers as shown in Fig.~\ref{fig:second framework}. The final fusion is a decision level fusion between the scores generated by two classifiers. We use maximum fusion strategy to generate final classification results.

\subsubsection{Computationally Efficient Fusion Framework}

The architecture of the \textit{computationally efficient fusion framework} is shown in Fig.~\ref{fig:third framework}.

\begin{figure}[h]
	\centering
	\includegraphics[width=0.9\linewidth]{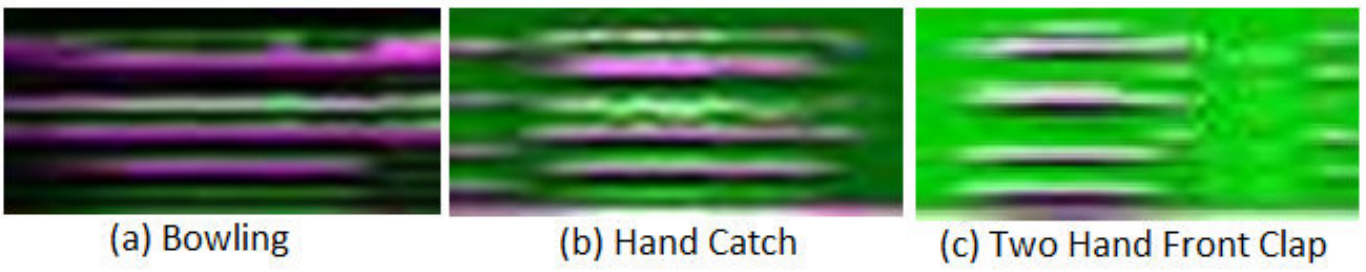}
	\caption{Composite Signal Images of three different actions}
	\label{fig:composite images}
\end{figure}

The first major change in this fusion framework is that only two CNNs are used as compared with first two frameworks where four CNNs are used, making it computationally far more efficient. Furthermore, the hybrid fusion in this framework is performed as combination of image fusion (data level fusion) and feature level fusion as compared to deep hybrid fusion framework where hybrid fusion was the combination of feature level and decision level fusions. We perform image level fusion between signal images and Prewitt filtered signal images to generate composite images as shown in Fig.~\ref{fig:composite images}.

\begin{table} 
	
	\hspace{-0.7cm}
	\begin{adjustbox}{width=\columnwidth,center}
		\renewcommand\arraystretch{1}
		\begin{tabular}{|c|c|c|c|c|}
			\hline
			\textbf{Dataset}     & \textbf{\makecell{Sampling \\ Rate}}    &\textbf{ Modality}    & \textbf{\makecell{Training\\Samples}}   & \textbf{\makecell{Test\\ Samples}}   \\
			\hline

			\multirow{3}*{UTD MHAD}
			& \multirow{2}*{50}
			& Depth    & 46639    & 11660    \\
			\cline{3-5}
			&           & \multirow{1}*{Inertial}
			& 11031    & 2745    \\\hline
			
			\multirow{3}*{Berkeley MHAD}
			& \multirow{2}*{30}
			& Depth    & 26400    & 6500    \\
			\cline{3-5}
			&           & \multirow{1}*{Inertial}
			& 2612    & 653    \\\hline
			
			\multirow{3}*{UTD Kinect V2}
			& \multirow{2}*{50}
			& Depth    & 14098    & 3524   \\
			\cline{3-5}
			&           & \multirow{1}*{Inertial}
			& 3532    & 884   \\\hline
			
		\end{tabular}
	\end{adjustbox}
	\caption{Dataset Information}
	\label{dataset information}
\end{table}

\begin{figure}
	\vspace{0.5cm}
	\hspace{-0.8cm}
	\centering
	\includegraphics[width=\linewidth]{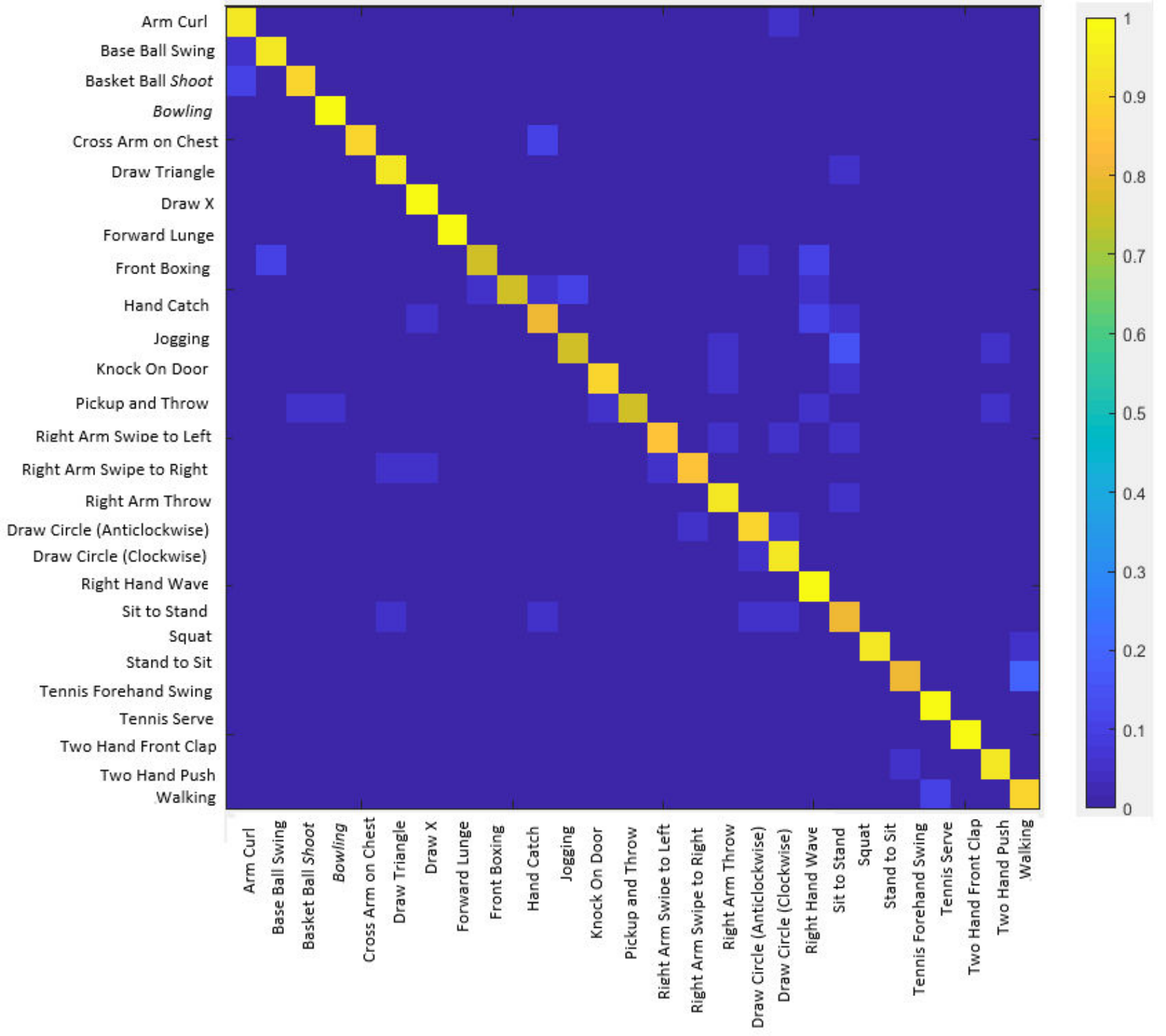}
	\caption{Confusion Matrix of UTD-MHAD Dataset for baseline experiment with LSTM. True labels are on vertical axis and  predicted labels are on horizontal axis}
	\label{fig:UTD-baseline-RNN}
\end{figure}
The composite image consists of different color channels. Gray region in the composite image shows the region where the two images have the same intensities. Green and magenta colors show the region where the intensities are different. Moreover, in this framework shown in Fig.~\ref{fig:third framework}, we did not make depth modality multimodal to keep this fusion framework computationally efficient. Alexnet is used to extract features from depth images and the CNN , whose architecture is shown in Fig.~\ref{fig:CNN Architecture}, is used to extract features from the composite images. The final fusion is CCF to get the most discriminant features from the input modalities. 

We used deep learning models in our proposed fusion framework. Deep learning models extract features at all levels of their structure and thus low level, high level and complex features are extracted. On the other hand non-deep learning models capture only the subset of features and thus have a generalization problem. Hence the higher accuracy is achieved with the deep models when compared to non-deep models and this can be seen in the comparison tables of the datasets. Although the improvement is marginal but at these higher accuracies, we believe even the marginal improvements are considerable. It means that our proposed frameworks can successfully classify even the difficult instances, where other methods are unsuccessful.

\begin{table}[h]
	\vspace{0.3cm}
	\centering
	\begin{tabular}{|c|c|}
		
		\hline 
		\textbf{Training Parameters} & \textbf{Values}   \\\hline 
		Momentum  &      0.9 \\\hline
		Initial Learn Rate  &      0.005 \\\hline
		Learn Rate Drop Factor  &      0.5 \\\hline
		Learn Rate Drop Period  &      10 \\\hline
		$L_2$ Regularization  &      0.004 \\\hline
		Max Epochs  &      50\\\hline
		MiniBatchSize  &   128 \\\hline		
		
	\end{tabular}
	\caption{Training Parameters for AlexNet}
	\label{tab:parameters for alexnet}
\end{table}

\begin{table}[h]
	\vspace{0.3cm}
	\centering
	\begin{tabular}{|c|c|}
		
		\hline 
		\textbf{Training Parameters} & \textbf{Values}   \\\hline 
		Momentum  &      0.9 \\\hline
		Initial Learn Rate  &      0.001 \\\hline
		Learn Rate Drop Factor  &      0.5 \\\hline
		Learn Rate Drop Period  &      10 \\\hline
		$L_2$ Regularization  &      0.004 \\\hline
		Max Epochs  &      100\\\hline
		MiniBatchSize  &   64 \\\hline		
		
	\end{tabular}
	\caption{Training Parameters for the CNN for signal iamges}
	\label{tab:parameters for CNN}
\end{table}

\section{Experiments and Results} \label{Experiment and results}
We experiment on three publicly available Multimodal Human Action Datasets, namely, UTD-MHAD~\cite{chen2015utd}, Berkeley MHAD~\cite{ofli2013berkeley} and UTD Kinect-V2 dataset~\cite{blog}. We used subject specific setting for experiments on all datasets. In subject specific setting, training and testing sets are split randomly across all subjects.

For our experiments on all deep $M^2$ fusion frameworks, we split the datasets into training and testing samples by randomly splitting 80\% data into training and 20\% data into testing samples. The number of training and testing samples after splitting are shown in Table~\ref{dataset information}. We ran the random split 20 times and report the average accuracy. We fine tune AlexNet on the SFIs obtained from the depth sequences and Prewitt filtered SFIs for 50 epochs. The values of other training parameters are shown in Table~\ref{tab:parameters for alexnet}. We reached these values through using the grid search method for hyperparameter tuning. 

Convolutional neural network, whose architecture is shown in Fig.~\ref{fig:CNN Architecture}, is trained on signal images and Prewitt filtered signal images with parameters shown in Table~\ref{tab:parameters for CNN}. The AlexNet and CNN described above are used in experiments on all three deep $M^2$ fusion frameworks explained in section~\ref{proposed method}. We conduct our experiments on Matlab R2018b on a desktop computer with NVIDIA GTX-1070 GPU.

\begin{figure}
	\hspace{-0.5cm}
	\centering
	\includegraphics[width=\linewidth]{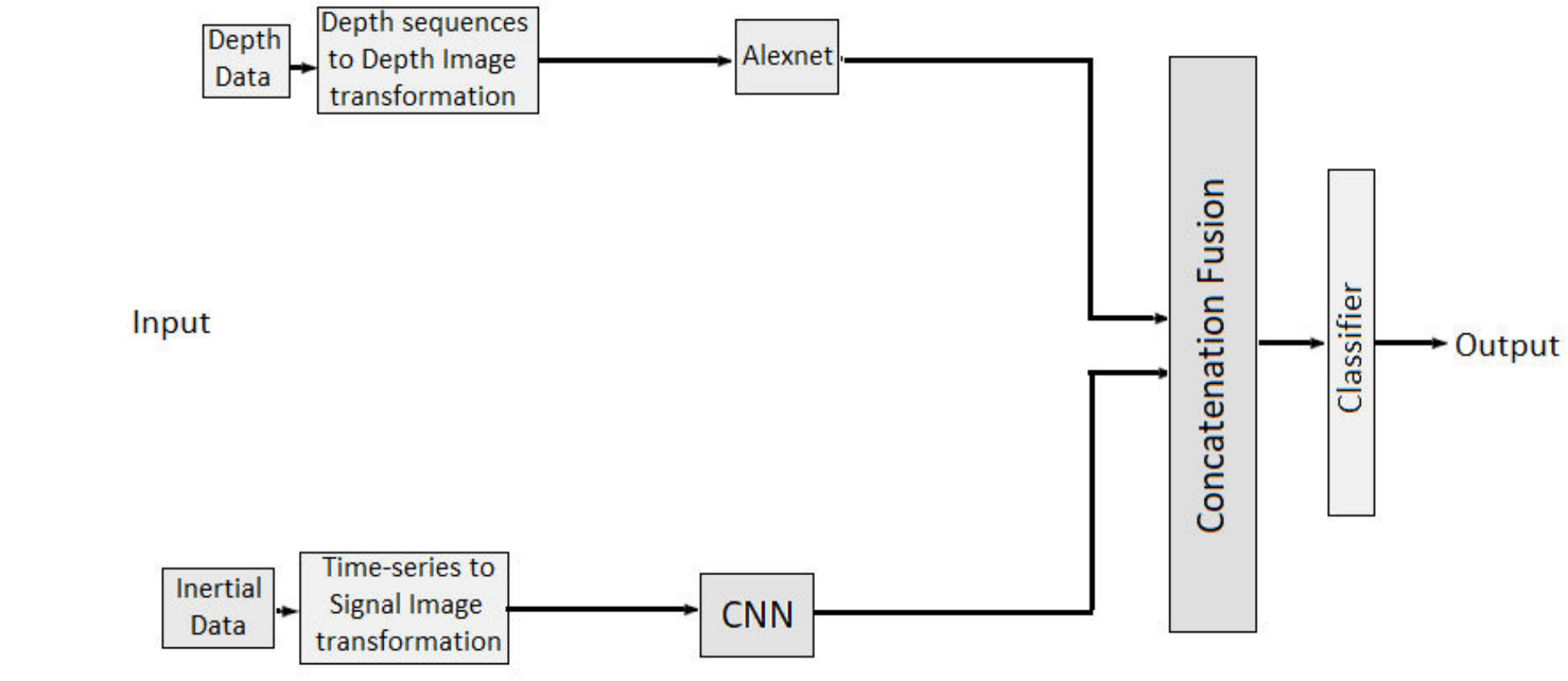}
	
	\caption{Fusion Framework for baseline Experiments}
	\label{fig: baseline framework}
\end{figure}

\subsection{UTD-MHAD Dataset}\label{MHAD dataset}
The UTD-MHAD dataset was collected in an indoor environment and contains both depth and inertial data components. The names of these 27 different actions are shown along the vertical and horizontal axis of Fig.~\ref{fig:UTD-baseline-RNN}.

 The inertial sensor component of UTD-MHAD dataset is very challenging to train a CNN. The first deficiency is that inertial sensor was worn either on volunteer's right wrist or right thigh depending upon nature of action. Hence the  sensor is worn only on two positions for collecting data of 27 actions which is not enough to capture all the dependencies and characteristics of data. The other challenge is that the number of data samples is very small for training a deep network.

To overcome these problems, we perform data augmentation on signal images of each dataset to increase the number of training and testing samples by applying the data augmentation techniques discussed in~\cite{ahmad2018towards}. The number of samples shown in Table~\ref{dataset information} are obtained after augmentation of signal images.

We perform experiments with all three deep $M^2$ fusion frameworks using UTD-MHAD datsets with same number of samples shown in Table~\ref{dataset information} and with same parameters of AlexNet and CNN shown in Tables~\ref{tab:parameters for alexnet} and~\ref{tab:parameters for CNN}. The results in terms of recognition accuracies and their comparison with previous state of art are shown in Table~\ref{tab:UTDMHAD comparisonTabe}. 

\subsubsection{Baseline Experiments}

We performed baseline experiments with UTD-MHAD dataset to validate the effect of our proposed signal to image transformation scheme. For these experiments, the same number of training and testing samples shown in Table~\ref{dataset information} were used. We first trained Long Short-Term Memory (LSTM) Network, a variant of recurrent neural network (RNN) that can learn long-term dependencies between time steps of sequence data, on raw inertial data only as LSTM is a well established  deep model for sequential data~\cite{greff2016lstm}. The architecture of an LSTM used in our experiment consists of 200 hidden layers, 27 fully connected layers followed by a softmax and classification layer. We trained LSTM for 200 epochs and obtained an accuracy of 86\%. The confusion matrix obtained by training LSTM with raw inertial data is shown in  Fig.~\ref{fig:UTD-baseline-RNN}. The blue patches on both sides of the diagonal show the misclassification results.

In order to improve accuracy we employed ID CNN on raw inertial multivariate time series. The input is a matrix of 52 time steps times 6 features.
50 kernels of size 5 x 1 are used in first convolutional layer, followed by 2 x 1 subsampling layer. The second convolutional layer contains 100 filters of same size followed by 2 x 1 subsampling layer, a fully connected layer and a classification layer. We obtained an accuracy of 77\% as shown in Table~\ref{tab:Baseline}.

 This poor performance of raw inertial data with LSTM and 1D CNN compelled us to transform the raw inertial data into more informative form. Thus to improve baseline experiments, we transform raw inertial data into signal images and depth data into depth images and employed CNNs to extract features as CNN is typically designed to perform well on images. 
 
 We first employed AlexNet on signal images. For this we did preprocessing and convert the original 24 x 52 signal images into 227 x 227. We obtained accuracy of 91.8\%. However by using our own designed CNN, whose architecture is shown in Fig.~\ref{fig:CNN Architecture}, we achieved recognition accuracy of 93.7\%. Since Alexnet is a deeper network compared to our CNN architecture, we suspect the reason for this could be AlexNet overfitting on the training data. 
 
 The fusion framework for performing baseline experiments is same as of our recent work~\cite{ahmad2018towards} and is shown in Fig~\ref{fig: baseline framework}. The results for baseline experiments are shown in Table~\ref{tab:Baseline}. The effect of signal to image transformation is significant as the accuracy of only inertial data (in the form of signal images) increases nearly 8\%. Furthermore using simple concatenation fusion between inertial and depth modalities the overall recognition accuracy rises to 98.4\%. The performance of simple fusion framework, where fusion is performed at single stage, encouraged us to design the proposed $M^2$ fusion frameworks. As we can see in Table~\ref{tab:UTDMHAD comparisonTabe}, the proposed $M^2$ fusion frameworks outperform our baseline that was presented in~\cite{ahmad2018towards}.
 
 The inference speed of all three proposed fusion frameworks is calculated and is shown in Table~\ref{tab:computational speed}. Inference speed is expressed in microseconds ($\mu$s). It is defined as a time
 taken by classifier to classify each test sample. 
 
 Further disscussion on results can be found in section~\ref{discussion on results}.

\begin{figure*}
	\centering
	\includegraphics[width=0.7\linewidth]{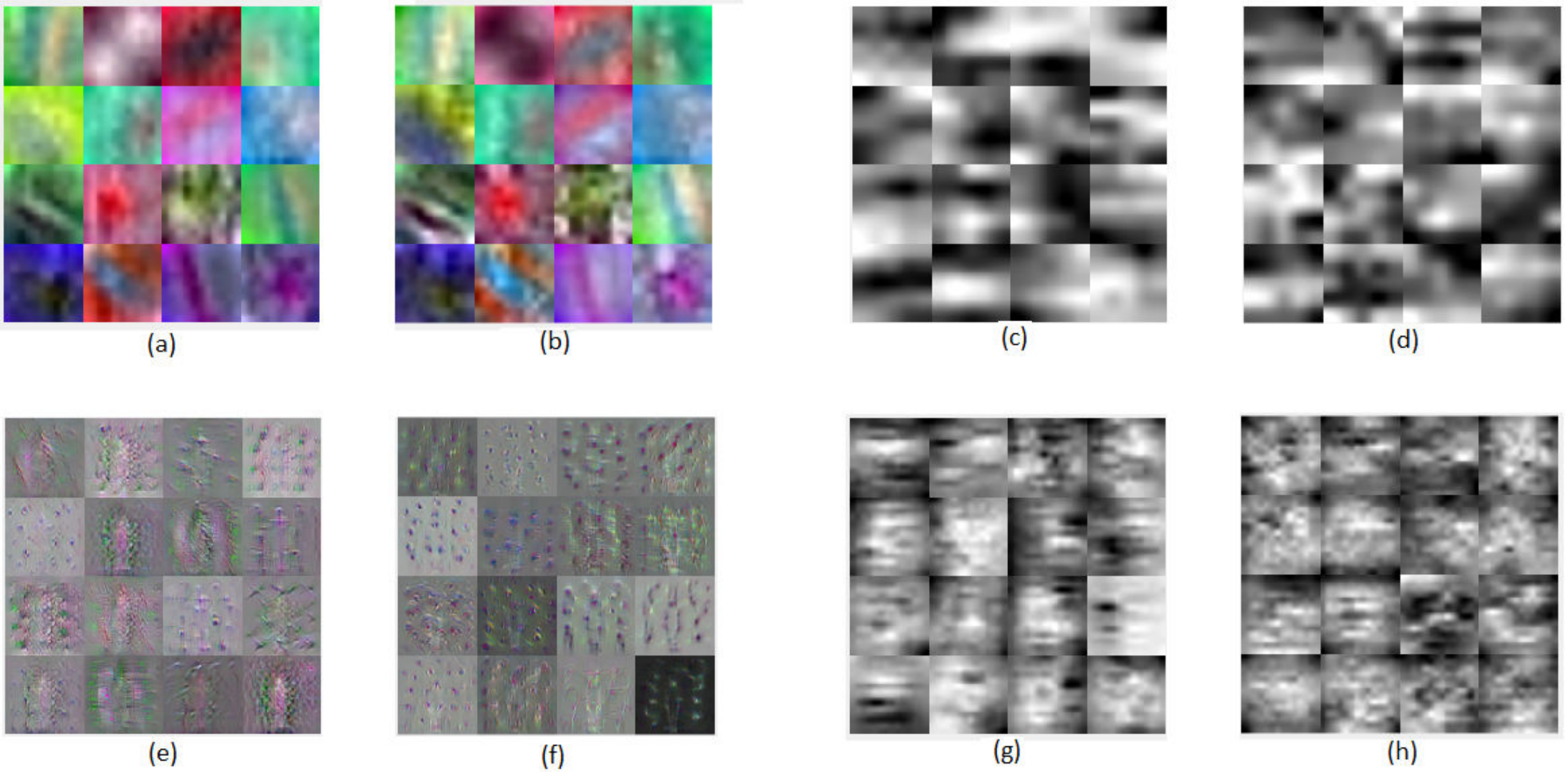}
	\caption{Feature visualization from first convolutional layer. (a) Alexnet on SFIs (b) Alexnet on Prewitt filtered SFIs (c) CNN on signal images (d) CNN on Prewitt filtered signal images. Feature visualization from last convolutional layer. (e) Alexnet on SFIs (f) Alexnet on Prewitt filtered SFIs (g) CNN on signal images (h) CNN on Prewitt filtered signal images.}
	\label{fig:Feature visualization}
\end{figure*}
\begin{table}[h]
\begin{adjustbox}{width=\columnwidth,center}
	\begin{tabular}{|c|c|}
		
		\hline 
		\textbf{Previous Methods} & \textbf{Accuracy\%}  \\\hline 
		C.chen et al.~\cite{chen2015utd}       &      97.1 \\\hline
		Bulbul et al.~\cite{bulbul2015dmms}    &      88.4 \\\hline
		N.Dawar et al. ~\cite{dawar2019data}             &      89.2 \\\hline
		N.Dawar et al. ~\cite{dawar2018action}            &      92.8 \\\hline
		Chen et al.~\cite{chen2016real}        &      97.2 \\\hline
		Mahjoub et al.~\cite{mahjoub2018efficient}        &      98.5 \\\hline
		Our baseline ~\cite{ahmad2018towards}              &      98.4 \\\hline

		\multicolumn{2}{|l|}{\textbf{\textit{Performance of our proposed fusion frameworks}}}\\\hline
		
		\multicolumn{2}{|l|}{\textbf{1. Deep Multistage Feature Fusion Framework}}\\\hline
		SFI + Signal image (SI) & 98.5 \\\hline
		SFI + SI + Prewitt filtered SI & 98.6 \\\hline
	    SFI + Prewitt filtered SFI + Signal image & 98.9 \\\hline
    	SFI + Prewitt filtered SFI + SI + Prewitt filtered SI & 99 \\\hline
    	\multicolumn{2}{|l|}{\textbf{2. Deep Hybrid Fusion Framework}}\\\hline
    	SFI + Signal image (SI) & 99 \\\hline
    	SFI + SI + Prewitt filtered SI & 99.08 \\\hline
    	SFI + Prewitt filtered SFI + Signal image & 99.2 \\\hline
    	SFI + Prewitt filtered SFI + SI + Prewitt filtered SI & 99.3 \\\hline
	
		\makecell{\textbf{3. Computationally Efficient Fusion Framework} }  & 99.2  \\\hline
		
	\end{tabular}
\end{adjustbox}
	\caption{ Comparison of Accuracies of Deep  $M^2$ Fusion Frameworks with previous methods on UTD-MHAD dataset}
	\label{tab:UTDMHAD comparisonTabe}
\end{table}

\begin{table}[h]
	
	\begin{adjustbox}{width=\columnwidth,center}
	\begin{tabular}{|c|c|}
		
		\hline 
		\textbf{Fusion Frameworks}& \textbf{Inference Speed ($\mu$s)}  \\\hline 
		Deep Multistage Feature Fusion Framework & 158  \\\hline	
		Deep Hybrid Fusion Framework & 235  \\\hline
		Computationally Efficient Fusion Framework & 147\\\hline
		
	\end{tabular}
\end{adjustbox}
	\caption{Inference Speed of Proposed Fusion Framework for UTD-MHAD Dataset}
	\label{tab:computational speed}
\end{table}

\begin{table}[h]
	\centering
	\begin{tabular}{|c|c|}	
		\hline 
		\textbf{Methods} & \textbf{Accuracy\%}   \\\hline 
		Raw Inertial data with RNN  &      86 \\\hline
		Raw Inertial data with 1D CNN & 77    \\\hline
		Signal Images (SI) with AlexNet    & 91.8 \\\hline
		Signal Images with CNN of Fig.~\ref{fig:CNN Architecture}  &      93.7 \\\hline
		Front view Images with AlexNet  &      96.1 \\\hline
		Fusion of Front, Side and Top view Images & 96.8 \\\hline 
		Baseline fusion~\cite{ahmad2018towards}   &      98.4 \\\hline	
	    Decision level fusion (Depth + Inertial)   &      98.6 \\\hline
	\end{tabular}
	\caption{Baseline Experiments on UTD-MHAD Dataset}
	\label{tab:Baseline}
\end{table}

\subsection{Berkeley Multimodal Human Action Dataset}
The Berkeley MHAD contains 11 actions performed five times by seven male and five female subjects. 

 There are five modalities in the dataset from which we used depth and inertial sensor modalities for experiments as the combination of these to modalities is cost effective and easy to handle.

 Inertial part of the dataset contains six accelerometers and each generates three sequences. For generating signal images explained in section~\ref{signal images}, we need six sequences in a row. Thus we used two accelerometer($A_1$ and $A_4$) and stacked them row wise to make six sequences. The reason for selecting $A_1$ and $A_4$ is that they are worn on the left wrist and right hip, respectively and are able to generate more useful information than those worn on both ankles~\cite{chen2015improving}.
 We performed experiments on all three deep $M^2$ fusion frameworks using Berkeley MHAD dataset with same settings described in Table~\ref{dataset information},~\ref{tab:parameters for alexnet} and~\ref{tab:parameters for CNN}.
 The results and their comparison with previous state of art are shown in Table~\ref{tab:results on Berkeley MHAD dataset}. Results are analyzed in section~\ref{discussion on results}.
 \begin{table}[h]
 \begin{adjustbox}{width=\columnwidth,center}
 	\begin{tabular}{|c|c|}
 		
 		\hline 
 		\textbf{Previous Methods} & \textbf{Accuracy\%}  \\\hline 
 		F. Ofli et al.~\cite{ofli2013berkeley}       &      97.81 \\\hline
 		Alireza Shafaei et al.~\cite{shafaei2016real} &     98.1   \\\hline
 		Earnest Paul Ijjina et al.~\cite{ijjina2014human} &  98.38 \\\hline
 		Chen Chen et al.~\cite{chen2015improving}       &      99.54 \\\hline
 		\multicolumn{2}{|l|}{\textbf{\textit{Deep Multilevel Multimodal Fusion Frameworks}}}\\
 		\hline
 		Deep Multistage Feature Fusion Framework & 99.4 \\\hline
 		Deep Hybrid Fusion Framework  & 99.8 \\\hline
 		\makecell{Computationally Efficient Fusion Framework}   & 99.6  \\\hline			
 	\end{tabular}
 \end{adjustbox}
 	\caption{Comparison of Accuracies of Deep  $M^2$ Fusion Frameworks with previous methods on Berkeley MHAD dataset}
 	\label{tab:results on Berkeley MHAD dataset}
 \end{table}
 \begin{table}[h]
	\begin{adjustbox}{width=\columnwidth,center}
	\begin{tabular}{|c|c|}
		\hline 
		\textbf{Previous Methods} & \textbf{Accuracy\%}  \\\hline 
		Chen et al.~\cite{chen2016fusion}       &      99.5 \\\hline
		Z.Ahmad et al.~\cite{ahmad2018towards}  &      99.8  \\\hline
		\multicolumn{2}{|l|}{\textbf{\textit{Deep Multilevel Multimodal Fusion Frameworks}}}\\
		\hline
		Deep Mulitstage Feature Fusion Framework & 99.6 \\\hline
		Deep Hybrid Fusion Framework  & 99.8 \\\hline
		\makecell{Computationally Efficient Fusion Framework}   & 99.8  \\\hline			
	\end{tabular}
\end{adjustbox}
	\caption{Comparison of Accuracies of Deep $M^2$ Fusion Frameworks with previous methods on Kinect V2 dataset}
	\label{tab:results on Kinect 2 dataset}
\end{table}

\subsection{UTD kinect V2 Dataset}\label{Kinect2 dataset}

KinectV2 action dataset is another publicly available dataset that contains both depth and inertial data. It is a new dataset using the second generation of kinect. It contains 10 actions performed by six subjects with each subject repeating the action 5 times.

 We performed experiments on all three deep $M^2$ fusion frameworks using Kinect V2 dataset with same settings described in Table~\ref{dataset information},~\ref{tab:parameters for alexnet} and~\ref{tab:parameters for CNN}.
 The results in terms of recognition accuracies and their comparison with previous state of art are shown in Table~\ref{tab:results on Kinect 2 dataset}.

\let\thefootnote\relax\footnote{The preprocessed dataset and related code can be found at https://github.com/zaamad/Deep-Multilevel-Multimodal-Fusion}

\subsection{Discussion on Results}\label{discussion on results}

 Experimental results on proposed deep $M^2$ fusion frameworks are shown in Tables~\ref{tab:UTDMHAD comparisonTabe},~\ref{tab:results on Berkeley MHAD dataset} and~\ref{tab:results on Kinect 2 dataset}. These results show the consistency and state of art performance of the proposed fusion frameworks described in section~\ref{proposed method}. The uniform performance of all deep $M^2$ fusion frameworks is obvious from the fact that all yield high accuracy for Kinect V2 dataset and lower accuracy for UTD-MHAD dataset. This is due to the fact that in UTD-MHAD there are actions which are less discriminative such as "sit to stand" and "stand to sit" and "right arm swipe to left" and "right arm swipe to right". Furthermore, interclass discrimination in Kinect V2 dataset is higher than UTD-MHAD and Berkeley MHAD datasets. 
  
 The objective of making input modalities further multimodal by convolving with Prewitt filter, played significant role in the state of art performance of the proposed fusion frameworks. Prewitt filtering intensify the edges or sharp changes in the images and allow CNN to extract more discriminative features alongside unfiltered images and thus enhance the performance of proposed frameworks. These discriminative and complementary features extracted from the first and last convolutional layers of AlexNets and CNNs can be visualized in Fig.~\ref{fig:Feature visualization}. For visualization and to observe the discriminative and complementary nature of features at different levels of AlexNet and CNN, we extracted 16 features from each of the first and last convolutional layers of AlexNet and CNN shown in the constituent figures of Fig.~\ref{fig:Feature visualization}. The distinguishable orientation, brightness, intensity and other related features such as color, shown in these 16 extracted features reflect the discriminative and complemenatary nature of the features at different levels of AlexNet and CNN and thus consolidate the idea of creating modality within the modality to improve the performance of the deep $M^2$ fusion frameworks.

\section{Conclusion}

In this paper, we present three novel deep multilevel multimodal ($M^2$) fusion frame works for improving the accuracy of Human Action Recognition (HAR) using depth and inertial modalities.	At the input of each fusion framework, we transform the depth data and inertial sensor data into Sequential Front View Images (SFI) and Signal Images (SI) respectively. We made input modalities, depth and inertial, further multimodal by taking convolution with an edge detector called Prewitt filter. We extract distinct and complementary features from the input modalities by employing CNNs and fused these features at more than one stages in our three novel fusion frameworks. Rich features obtained after multilevel fusion are inputted to the SVM classifier. The state of the art recognition accuracy achieved by our three fusion frameworks on three publicly available multimodal human action datasets show the dominance of the proposed deep  $M^2$ fusion frameworks. 

	 
 \bibliographystyle{IEEEtran}

\begin{thebibliography}{10}
 	\providecommand{\url}[1]{#1}
 	\csname url@samestyle\endcsname
 	\providecommand{\newblock}{\relax}
 	\providecommand{\bibinfo}[2]{#2}
 	\providecommand{\BIBentrySTDinterwordspacing}{\spaceskip=0pt\relax}
 	\providecommand{\BIBentryALTinterwordstretchfactor}{4}
 	\providecommand{\BIBentryALTinterwordspacing}{\spaceskip=\fontdimen2\font plus
 		\BIBentryALTinterwordstretchfactor\fontdimen3\font minus
 		\fontdimen4\font\relax}
 	\providecommand{\BIBforeignlanguage}[2]{{%
 			\expandafter\ifx\csname l@#1\endcsname\relax
 			\typeout{** WARNING: IEEEtran.bst: No hyphenation pattern has been}%
 			\typeout{** loaded for the language `#1'. Using the pattern for}%
 			\typeout{** the default language instead.}%
 			\else
 			\language=\csname l@#1\endcsname
 			\fi
 			#2}}
 	\providecommand{\BIBdecl}{\relax}
 	\BIBdecl
 	
 	\bibitem{chen2014medication}
 	C.~Chen, N.~Kehtarnavaz, and R.~Jafari, ``A medication adherence monitoring
 	system for pill bottles based on a wearable inertial sensor,'' in \emph{2014
 		36th Annual International Conference of the IEEE Engineering in Medicine and
 		Biology Society}.\hskip 1em plus 0.5em minus 0.4em\relax IEEE, 2014, pp.
 	4983--4986.
 	
 	\bibitem{corbishley2008breathing}
 	P.~Corbishley and E.~Rodriguez-Villegas, ``Breathing detection: towards a
 	miniaturized, wearable, battery-operated monitoring system,'' \emph{IEEE
 		Transactions on Biomedical Engineering}, vol.~55, no.~1, pp. 196--204, 2008.
 	
 	\bibitem{zhou2016never}
 	B.~Zhou, M.~Sundholm, J.~Cheng, H.~Cruz, and P.~Lukowicz, ``Never skip leg day:
 	A novel wearable approach to monitoring gym leg exercises,'' in \emph{2016
 		IEEE International Conference on Pervasive Computing and Communications
 		(PerCom)}.\hskip 1em plus 0.5em minus 0.4em\relax IEEE, 2016, pp. 1--9.
 	
 	\bibitem{qin2016compressive}
 	J.~Qin, L.~Liu, Z.~Zhang, Y.~Wang, and L.~Shao, ``Compressive sequential
 	learning for action similarity labeling,'' \emph{IEEE Transactions on Image
 		Processing}, vol.~25, no.~2, pp. 756--769, 2016.
 	
 	\bibitem{plotz2011feature}
 	T.~Pl{\"o}tz, N.~Y. Hammerla, and P.~L. Olivier, ``Feature learning for
 	activity recognition in ubiquitous computing,'' in \emph{Twenty-Second
 		International Joint Conference on Artificial Intelligence}, 2011.
 	
 	\bibitem{krizhevsky2012imagenet}
 	A.~Krizhevsky, I.~Sutskever, and G.~E. Hinton, ``Imagenet classification with
 	deep convolutional neural networks,'' in \emph{Advances in neural information
 		processing systems}, 2012, pp. 1097--1105.
 	
 	\bibitem{aggarwal2014human}
 	J.~K. Aggarwal and L.~Xia, ``Human activity recognition from 3d data: A
 	review,'' \emph{Pattern Recognition Letters}, vol.~48, pp. 70--80, 2014.
 	
 	\bibitem{yang2009distributed}
 	A.~Y. Yang, R.~Jafari, S.~S. Sastry, and R.~Bajcsy, ``Distributed recognition
 	of human actions using wearable motion sensor networks,'' \emph{Journal of
 		Ambient Intelligence and Smart Environments}, vol.~1, no.~2, pp. 103--115,
 	2009.
 	
 	\bibitem{chen2017survey}
 	C.~Chen, R.~Jafari, and N.~Kehtarnavaz, ``A survey of depth and inertial sensor
 	fusion for human action recognition,'' \emph{Multimedia Tools and
 		Applications}, vol.~76, no.~3, pp. 4405--4425, 2017.
 	
 	\bibitem{ramachandram2017deep}
 	D.~Ramachandram and G.~W. Taylor, ``Deep multimodal learning: A survey on
 	recent advances and trends,'' \emph{IEEE Signal Processing Magazine},
 	vol.~34, no.~6, pp. 96--108, 2017.
 	
 	\bibitem{hall1997introduction}
 	D.~L. Hall and J.~Llinas, ``An introduction to multisensor data fusion,''
 	\emph{Proceedings of the IEEE}, vol.~85, no.~1, pp. 6--23, 1997.
 	
 	\bibitem{wu2006multi}
 	Z.~Wu, L.~Cai, and H.~Meng, ``Multi-level fusion of audio and visual features
 	for speaker identification,'' in \emph{International Conference on
 		Biometrics}.\hskip 1em plus 0.5em minus 0.4em\relax Springer, 2006, pp.
 	493--499.
 	
 	\bibitem{atrey2007goal}
 	P.~K. Atrey, M.~S. Kankanhalli, and J.~B. Oommen, ``Goal-oriented optimal
 	subset selection of correlated multimedia streams,'' \emph{ACM Transactions
 		on Multimedia Computing, Communications, and Applications (TOMM)}, vol.~3,
 	no.~1, p.~2, 2007.
 	
 	\bibitem{ni2004image}
 	J.~Ni, X.~Ma, L.~Xu, and J.~Wang, ``An image recognition method based on
 	multiple bp neural networks fusion,'' in \emph{International Conference on
 		Information Acquisition, 2004. Proceedings.}\hskip 1em plus 0.5em minus
 	0.4em\relax IEEE, 2004, pp. 323--326.
 	
 	\bibitem{hatami2018classification}
 	N.~Hatami, Y.~Gavet, and J.~Debayle, ``Classification of time-series images
 	using deep convolutional neural networks,'' in \emph{Tenth International
 		Conference on Machine Vision (ICMV 2017)}, vol. 10696.\hskip 1em plus 0.5em
 	minus 0.4em\relax International Society for Optics and Photonics, 2018, p.
 	106960Y.
 	
 	\bibitem{chen2015improving}
 	C.~Chen, R.~Jafari, and N.~Kehtarnavaz, ``Improving human action recognition
 	using fusion of depth camera and inertial sensors,'' \emph{IEEE Transactions
 		on Human-Machine Systems}, vol.~45, no.~1, pp. 51--61, 2015.
 	
 	\bibitem{manzi2018enhancing}
 	A.~Manzi, A.~Moschetti, R.~Limosani, L.~Fiorini, and F.~Cavallo, ``Enhancing
 	activity recognition of self-localized robot through depth camera and
 	wearable sensors,'' \emph{IEEE Sensors Journal}, vol.~18, no.~22, pp.
 	9324--9331, 2018.
 	
 	\bibitem{chen2016real}
 	C.~Chen, R.~Jafari, and N.~Kehtarnavaz, ``A real-time human action recognition
 	system using depth and inertial sensor fusion,'' \emph{IEEE Sensors Journal},
 	vol.~16, no.~3, pp. 773--781, 2016.
 	
 	\bibitem{liu2014fusion}
 	K.~Liu, C.~Chen, R.~Jafari, and N.~Kehtarnavaz, ``Fusion of inertial and depth
 	sensor data for robust hand gesture recognition,'' \emph{IEEE Sensors
 		Journal}, vol.~14, no.~6, pp. 1898--1903, 2014.
 	
 	\bibitem{dawar2018real}
 	N.~Dawar and N.~Kehtarnavaz, ``Real-time continuous detection and recognition
 	of subject-specific smart tv gestures via fusion of depth and inertial
 	sensing,'' \emph{IEEE Access}, vol.~6, pp. 7019--7028, 2018.
 	
 	\bibitem{hu2018novel}
 	B.~H. Hu, N.~E. Krausz, and L.~J. Hargrove, ``A novel method for bilateral gait
 	segmentation using a single thigh-mounted depth sensor and imu,'' in
 	\emph{2018 7th IEEE International Conference on Biomedical Robotics and
 		Biomechatronics (Biorob)}.\hskip 1em plus 0.5em minus 0.4em\relax IEEE, 2018,
 	pp. 807--812.
 	
 	\bibitem{guo2017multiview}
 	Y.~Guo, D.~Tao, W.~Liu, and J.~Cheng, ``Multiview cauchy estimator feature
 	embedding for depth and inertial sensor-based human action recognition,''
 	\emph{IEEE Transactions on Systems, Man, and Cybernetics: Systems}, vol.~47,
 	no.~4, pp. 617--627, 2017.
 	
 	\bibitem{bernal2018deep}
 	E.~A. Bernal, X.~Yang, Q.~Li, J.~Kumar, S.~Madhvanath, P.~Ramesh, and R.~Bala,
 	``Deep temporal multimodal fusion for medical procedure monitoring using
 	wearable sensors,'' \emph{IEEE Transactions on Multimedia}, vol.~20, no.~1,
 	pp. 107--118, 2018.
 	
 	\bibitem{hwang2017multi}
 	I.~Hwang, G.~Cha, and S.~Oh, ``Multi-modal human action recognition using deep
 	neural networks fusing image and inertial sensor data,'' in \emph{2017 IEEE
 		International Conference on Multisensor Fusion and Integration for
 		Intelligent Systems (MFI)}.\hskip 1em plus 0.5em minus 0.4em\relax IEEE,
 	2017, pp. 278--283.
 	
 	\bibitem{dawar2019data}
 	N.~Dawar, S.~Ostadabbas, and N.~Kehtarnavaz, ``Data augmentation in deep
 	learning-based fusion of depth and inertial sensing for action recognition,''
 	\emph{IEEE Sensors Letters}, vol.~3, no.~1, pp. 1--4, 2019.
 	
 	\bibitem{dawar2018action}
 	N.~Dawar and N.~Kehtarnavaz, ``Action detection and recognition in continuous
 	action streams by deep learning-based sensing fusion,'' \emph{IEEE Sensors
 		Journal}, vol.~18, no.~23, pp. 9660--9668, 2018.
 	
 	\bibitem{ahmad2018towards}
 	Z.~Ahmad and N.~Khan, ``Towards improved human action recognition using
 	convolutional neural networks and multimodal fusion of depth and inertial
 	sensor data,'' in \emph{2018 IEEE International Symposium on Multimedia
 		(ISM)}.\hskip 1em plus 0.5em minus 0.4em\relax IEEE, 2018, pp. 223--230.
 	
 	\bibitem{dawar2018convolutional}
 	N.~Dawar and N.~Kehtarnavaz, ``A convolutional neural network-based sensor
 	fusion system for monitoring transition movements in healthcare
 	applications,'' in \emph{2018 IEEE 14th International Conference on Control
 		and Automation (ICCA)}.\hskip 1em plus 0.5em minus 0.4em\relax IEEE, 2018,
 	pp. 482--485.
 	
 	\bibitem{bobick2001recognition}
 	A.~F. Bobick and J.~W. Davis, ``The recognition of human movement using
 	temporal templates,'' \emph{IEEE Transactions on pattern analysis and machine
 		intelligence}, vol.~23, no.~3, pp. 257--267, 2001.
 	
 	\bibitem{jiang2015human}
 	W.~Jiang and Z.~Yin, ``Human activity recognition using wearable sensors by
 	deep convolutional neural networks,'' in \emph{Proceedings of the 23rd ACM
 		international conference on Multimedia}.\hskip 1em plus 0.5em minus
 	0.4em\relax Acm, 2015, pp. 1307--1310.
 	
 	\bibitem{gonzalez2002digital}
 	R.~C. Gonzalez, R.~E. Woods \emph{et~al.}, ``Digital image processing,'' 2002.
 	
 	\bibitem{uurtio2018tutorial}
 	V.~Uurtio, J.~M. Monteiro, J.~Kandola, J.~Shawe-Taylor, D.~Fernandez-Reyes, and
 	J.~Rousu, ``A tutorial on canonical correlation methods,'' \emph{ACM
 		Computing Surveys (CSUR)}, vol.~50, no.~6, p.~95, 2018.
 	
 	\bibitem{sun2005new}
 	Q.-S. Sun, S.-G. Zeng, Y.~Liu, P.-A. Heng, and D.-S. Xia, ``A new method of
 	feature fusion and its application in image recognition,'' \emph{Pattern
 		Recognition}, vol.~38, no.~12, pp. 2437--2448, 2005.
 	
 	\bibitem{chen2015utd}
 	C.~Chen, R.~Jafari, and N.~Kehtarnavaz, ``Utd-mhad: A multimodal dataset for
 	human action recognition utilizing a depth camera and a wearable inertial
 	sensor,'' in \emph{2015 IEEE International conference on image processing
 		(ICIP)}.\hskip 1em plus 0.5em minus 0.4em\relax IEEE, 2015, pp. 168--172.
 	
 	\bibitem{ofli2013berkeley}
 	F.~Ofli, R.~Chaudhry, G.~Kurillo, R.~Vidal, and R.~Bajcsy, ``Berkeley mhad: A
 	comprehensive multimodal human action database,'' in \emph{2013 IEEE Workshop
 		on Applications of Computer Vision (WACV)}.\hskip 1em plus 0.5em minus
 	0.4em\relax IEEE, 2013, pp. 53--60.
 	
 	\bibitem{blog}
 	\BIBentryALTinterwordspacing
 	Kinect2d dataset. [Online]. Available:
 	\url{http://www.utdallas.edu/~kehtar/Kinect2DatasetReadme.pdf}
 	\BIBentrySTDinterwordspacing
 	
 	\bibitem{greff2016lstm}
 	K.~Greff, R.~K. Srivastava, J.~Koutn{\'\i}k, B.~R. Steunebrink, and
 	J.~Schmidhuber, ``Lstm: A search space odyssey,'' \emph{IEEE transactions on
 		neural networks and learning systems}, vol.~28, no.~10, pp. 2222--2232, 2016.
 	
 	\bibitem{bulbul2015dmms}
 	M.~F. Bulbul, Y.~Jiang, and J.~Ma, ``Dmms-based multiple features fusion for
 	human action recognition,'' \emph{International Journal of Multimedia Data
 		Engineering and Management (IJMDEM)}, vol.~6, no.~4, pp. 23--39, 2015.
 	
 	\bibitem{mahjoub2018efficient}
 	A.~B. Mahjoub and M.~Atri, ``An efficient end-to-end deep learning architecture
 	for activity classification,'' \emph{Analog Integrated Circuits and Signal
 		Processing}, pp. 1--10, 2018.
 	
 	\bibitem{shafaei2016real}
 	A.~Shafaei and J.~J. Little, ``Real-time human motion capture with multiple
 	depth cameras,'' in \emph{2016 13th Conference on Computer and Robot Vision
 		(CRV)}.\hskip 1em plus 0.5em minus 0.4em\relax IEEE, 2016, pp. 24--31.
 	
 	\bibitem{ijjina2014human}
 	E.~P. Ijjina and C.~K. Mohan, ``Human action recognition based on mocap
 	information using convolution neural networks,'' in \emph{2014 13th
 		International Conference on Machine Learning and Applications}.\hskip 1em
 	plus 0.5em minus 0.4em\relax IEEE, 2014, pp. 159--164.
 	
 	\bibitem{chen2016fusion}
 	C.~Chen, R.~Jafari, and N.~Kehtarnavaz, ``Fusion of depth, skeleton, and
 	inertial data for human action recognition,'' in \emph{2016 IEEE
 		international conference on acoustics, speech and signal processing
 		(ICASSP)}.\hskip 1em plus 0.5em minus 0.4em\relax IEEE, 2016, pp. 2712--2716.
 	
 \end{thebibliography}
  
\end{document}